\documentclass[a4paper, 10pt, conference]{ieeeconf}      
\usepackage{FG2020}

\FGfinalcopy 

\IEEEoverridecommandlockouts                              
\def\FGPaperID{29} 

\title{\LARGE \bf
How are attributes expressed in face DCNNs?}

\author{\parbox{16cm}{\centering
    {\large Prithviraj Dhar$^1$, Ankan Bansal$^1$, Carlos D. Castillo$^1$, Joshua Gleason$^1$, P. Jonathon Phillips$^2$ and Rama Chellappa$^1$}\\
    {\normalsize
    $^1$ University of Maryland, College Park, MD, USA\\
    $^2$ National Institute of Standards and Technology, Gaithersburg, MD, USA}}%
}
\usepackage{tabularx}
\usepackage{algorithm}
\usepackage{algorithmic}
\usepackage[algo2e]{algorithm2e}
\usepackage{caption}
\usepackage{amssymb}
\usepackage{times}
\usepackage{epsfig}
\usepackage{graphicx}
\usepackage{amsmath}
\usepackage{amssymb}
\usepackage{siunitx}
\let\labelindent\relax
\usepackage{enumitem}
\usepackage{comment}
\usepackage{subfig}
\usepackage{booktabs}
\usepackage{makecell}
\usepackage{xcolor}
\usepackage{hyperref}
\usepackage{pifont}
\usepackage{xargs}                      
\usepackage[colorinlistoftodos,prependcaption,textsize=tiny]{todonotes}
\definecolor{red}{RGB}{255, 0, 0}
\definecolor{green}{RGB}{0, 255, 10}

\newcommand{\rulesep}{\unskip\ \vrule\ }
\newcommandx{\unsure}[2][1=]{\todo[linecolor=red,backgroundcolor=red!25,bordercolor=red,#1]{#2}}
\newcommandx{\change}[2][1=]{\todo[linecolor=blue,backgroundcolor=blue!25,bordercolor=blue,#1]{#2}}
\newcommandx{\info}[2][1=]{\todo[linecolor=OliveGreen,backgroundcolor=OliveGreen!25,bordercolor=OliveGreen,#1]{#2}}
\newcommandx{\improvement}[2][1=]{\todo[linecolor=gray,backgroundcolor=gray!25,bordercolor=gray,#1]{#2}}

\newlength\myindent
\begin{document}

\ifFGfinal
\thispagestyle{empty}
\pagestyle{empty}
\else
\author{Anonymous FG2020 submission\\ Paper ID \FGPaperID \\}
\pagestyle{plain}
\fi
\maketitle

\begin{abstract}
As deep networks become increasingly accurate at recognizing faces, it is vital to understand how these networks process faces. While these networks are solely trained to recognize identities, they also contain face related information such as sex, age, and pose of the face. The networks are not trained to learn these attributes.  We introduce expressivity as a measure of how much a feature vector informs us about an attribute, where a feature vector can be from internal or final layers of a network. Expressivity is computed by a second neural network whose inputs are features and attributes. The output of the second neural network  approximates the mutual information between feature vectors and an attribute. We investigate the expressivity for two different deep convolutional neural network (DCNN) architectures: a Resnet-101 and an Inception Resnet v2.  In the final fully connected layer of the networks, we found the order of  expressivity for facial attributes to be Age $>$ Sex $>$ Yaw. Additionally, we studied the changes in the encoding of facial attributes over training iterations. We found that as training progresses, expressivities of yaw, sex, and age decrease. Our technique can be a tool for investigating the sources of bias in a network and a step towards explaining the network's identity decisions.

\end{abstract}

\newcommand{\tcr}[1]{\textcolor{red}{#1}}
\newcommand{\pjp}[1]{\textcolor{red}{#1}}
\newcommand{\prithvi}[1]{\textcolor{blue}{#1}}

\section{Introduction}
\label{sec:intro}

Deep convolutional neural networks (DCNN)-based face algorithms are trained to learn the identity of a face; they are not trained to learn attributes of the face. These DCNNs generate representations that encode identity.  However, Hill \\emph{et al.} \cite{hill2019deepNatureMI} found that DCNNs generated identity representations self-organized by sex. Also, identity representations can contain information on pose, age, and illumination direction \cite{hill2019deepNatureMI,DBLP:journals/corr/abs-1904-01219,parde2017face}.

Facial attributes, including those mentioned above, affect algorithm accuracy \cite{Givens:2013fk,Lee:2014dn}. Assessing bias in algorithms implies measuring the effect of these attributes on accuracy.   Explaining how a network comes to an identity decision includes discerning how face representations encode attributes. To gain further understanding on the effects of attributes on bias and assist in developing methods to explain network decisions we address the following two questions.
How much information about  facial attributes are captured in the internal layers of the network?  How does the encoding of facial attributes evolve as training progresses?  

In this paper, we explore how attributes are encoded in the internal layers of two different and successful architectures: a Resnet-101 and an Inception Resnet v2 architecture based DCNN \cite{ranjan2019fast,bansal2018deep}. Both networks are trained solely to identify faces. In addition, we examine how the encoding of attributes evolves over training iterations for both these two networks.  To gain a greater understanding of the relationship between attributes, we introduce the concept of \textit{expressivity} of an attribute. Expressivity is a measure of how much a given representation informs us about an attribute, where  the representation of a face can be from internal or final layers in a DCNN. 
The following are the conceptual and experimental contributions of our paper.
\begin{enumerate}
    \item We are the first to investigate the encoding of facial attributes in the internal layers of DCNNs.
        \item For DCNNs, we are the first to monitor the evolution of the encoding of facial attributes during training.
    \item In the final fully connected layer of both networks, we observed that the order of  expressivity for three facial attributes to be Age $>$ Sex $>$ Yaw.
        \item We found that as training progresses, the expressivity of yaw, sex, and age decreases. The observed rate of decrease was Age $<$ Sex $<$ Yaw.
    \item The expressivity of identity dramatically increases from the last pooling layer to the final fully connected layer. 
\end{enumerate}

Knowing how face attributes are expressed in internal layers and how their representations evolve during training has both scientific and societal importance. Since DCNN-based face recognition systems are fielded in the real-world, the need for these networks to be explainable is pressing.  Understanding how internal layers encode attributes introduces new a tool to explain identity decisions and to examine the sources of bias in algorithms. 
From a scientific perspective, knowing the importance of attributes during training will provide insight into the training process.  
\section{Related work}
\label{sec:related}

Significant research has been done in training state-of-the-art face recognition networks  in the past few years \cite{ranjan2019fast,bansal2018deep,schroff2015facenet,taigman2014deepface,deng2018arcface}. Although, the interpretability of such networks has not been widely explored, several existing works explore explainability of deep networks for general visual recognition. These works can be divided into following two categories.\\

\textbf{Methods enforcing interpretability constraints during training}: Yin \emph{et al.} \cite{yin2018towards} propose a spatial activation diversity loss as a constraint to preserve interpretability while training face recognition networks. Similarly, Kim \emph{et al.} \cite{kim2014bayesian} propose a generative technique using the most representative exemplars (prototypes), thus highlighting interpretability of the model. As mentioned in \cite{kim2017interpretability}, these methods cannot be used to interpret pre-trained models and hence cannot be applied to networks or models which are already in use.\\

\textbf{Methods interpreting trained models} : TCAV \cite{kim2017interpretability} is one of the most influential work in this area. TCAV interprets a network on the basis of its sensitivity to user defined concepts (such as `stripes'). This is done by learning Concept Activity Vectors (CAVs) by training a linear classifier to distinguish between the activations produced by a concept's examples. While this method works efficiently for discrete physical concepts, such as presence of a specific color or pattern, it cannot be directly modified for checking the sensitivity of a model to a more general continuous concept (such as pose angle, age etc.). This is because for training CAVs, we also need negative example images where the concept being studied is missing. It is not trivial to find such images when the concepts are omnipotent and continuous (facial yaw, age etc.). Also, the method requires the testing images to belong to one of the training classes since the sensitivity computations requires measuring the change in logits of the class being investigated. This cannot be easily modified for our requirement where we use unseen subjects/faces to estimate models' sensitivity to facial attributes. Another important work in this category is \cite{alain2016understanding} where the authors use linear classifiers on different layers of a network to understand the role of intermediate layers. Koh and Liang  \cite{koh2017understanding} propose an influence function to measure the model's sensitivity to an infinitesimally-small local perturbation in the training images.  However, such a local perturbation-based method cannot be used to estimate models' sensitivity to physical attributes like pose or orientation. 

Another class of works \cite{selvaraju2017grad, chattopadhay2018grad} interprets the output of a network by generating saliency/attention maps. While such techniques help to highlight the spatial regions which affected the network's prediction, they do not allow to test the models' sensitivity to user defined concepts. Moreover, this method cannot be applied for concepts which cannot be physically localized (such as facial yaw, age etc.).

Hill \emph{et al.} \cite{hill2018deep} is one of the few works which interpret trained face recognition networks, where the authors show the following hierarchy: face identity nested under sex, illumination nested under identity, and viewpoint nested under illumination.

We interpret a trained face recognition network by investigating its sensitivity to facial attributes. Previous methods like \cite{koh2017understanding,kim2017interpretability} rely on the change in prediction with respect to a concept/attribute to interpret a network's sensitivity to an attribute. However, we introduce a new measure called expressivity, which quantifies the predictability of an attribute in a given set of features extracted using the model.  Moreover, expressivity can be computed for both categorical and continuous attributes, which enables us to compare the predictability of various attributes.

\section{Expressivity}
\label{sec:expressivity}
Predictability of facial attributes/identities in a given set of face descriptors indicates the attribute-relevant information content encoded in the descriptors. To estimate this information content, we intend to use Mutual Information (MI). MI between two random variables is a measure of the amount of information that can be obtained for one random variable by observing the other variable. Therefore, if we estimate MI between face descriptors and their corresponding identities/attributes, we can estimate the information content of these identities/attributes in the given descriptors. Since MI can be computed for both categorical and continuous attributes, an estimate based on MI  provides a  measure which is consistent across categorical and continuous attributes. 

MI between two random variables ($V_1, V_2$) is given as:
\begin{equation}
    I(V_1,V_2) = D_{KL}(\mathbb{P}_{V_1,V_2}\|\mathbb{P}_{V_1} \otimes \mathbb{P}_{V_2})
\end{equation}
where, $D_{KL}$ represents the Kullback-Leibler divergence, $\mathbb{P}_{V_1,V_2}$ denotes the joint probability distribution, $\mathbb{P}_{V_1}$ and $\mathbb{P}_{V_2}$ denote the marginal distributions, and $\mathbb{P}_{V_1} \otimes \mathbb{P}_{V_2}$ represents the product of the marginal distributions. Tishby and Zaslavsky \cite{tishby2015deep} show that each layer in a deep network can be quantified by the amount of mutual information (MI) it retains on the input variable, on the (desired) output variable. However, as mentioned in \cite{belghazi2018mine}, computing MI is not a trivial task. Most of the existing non-parametric approaches for estimating MI do not scale with the dimensionality of variables. Belghazi \emph{et al.} \cite{belghazi2018mine} propose MINE to estimate MI between high dimensional continuous variables using gradient descent over neural networks.
The neural information measure has been defined in \cite{belghazi2018mine} as follows.
\begin{equation}
    I_\Theta(F, A) = \underset{\theta \in \Theta}{\text{sup}} ~\mathbb{E}_{\mathbb{P}_{FA}}[T_\theta] - \text{log}(\mathbb{E}_{\mathbb{P}_{F}\otimes \mathbb{P_A}}[e^{T_\theta}])
\label{eq:mi}
\end{equation}
where $F, A$ are the variables whose mutual information is to be estimated, $\theta \in \Theta$ represents parameters in a network computing a function $T_\theta : F,A \longrightarrow \mathbb{R}$. As proved in \cite{belghazi2018mine}, the MINE estimate provides a lower bound estimate to the actual MI. 

In the context of our work, we define `Expressivity' of $A$ in $F$ as the aforementioned information measure (Equation \ref{eq:mi}). We use MINE to compute the expressivity of face identity and various facial attributes $A$ (discrete and continuous) in a given set of face descriptors $F$. Although face identity is not strictly a face `attribute', we treat it in the same way as a face attribute. Therefore, in this work, attribute $A$ collectively refers to identity and facial attributes like pose, sex etc.\\

\begin{figure}
 \centering
    \includegraphics[width=2in] {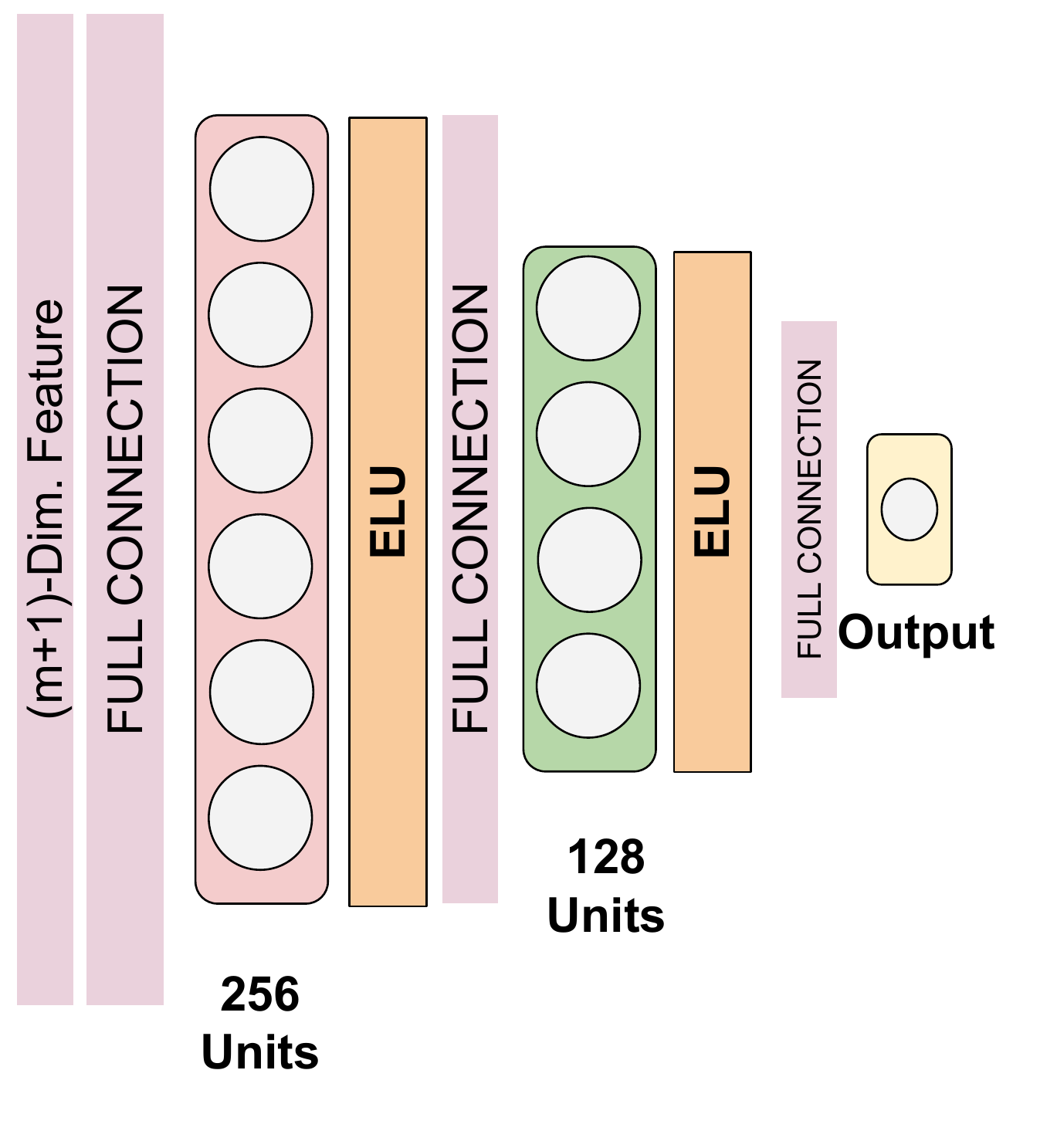}
    \caption{\small We use a consistent network architecture to compute expressivity of $m-$dimensional features $F$, augmented with an attribute vector $A$. More details in Sec. \ref{subsec:proto} }
    \label{fig:template}
    \vspace{-0.55cm}
\end{figure}

Following the protocols detailed in \cite{belghazi2018mine}, we briefly explain how gradient descent over a neural network can be used to compute expressivity. Let $f_i \in \mathbb{R}^m$ denote $i^{th}$ feature in a batch $B$ of size $b$ (i.e. $|B| = b$), and $a_i \in \mathbb{R}$ denote the corresponding attribute value. The set $\{(f_i, a_i)\}^{b}_{i=1}$ represents the $b$ elements sampled from the joint distribution ($f_i, a_i \sim \mathbb{P}_{FA}$). Similarly ${\{\Tilde{a_i}\}^{b}_{i=1}}$ represents $b$ attribute values sampled from a marginal distribution ($\Tilde{a_i} \sim \mathbb{P}_{A}$).  To estimate the the neural information in \ref{eq:mi}, we compute the expectation over joint and marginal distribution as follows :
$$\mathbb{E}_{\mathbb{P}^{(b)}_{FA}}[T_\theta] =\frac{1}{b}\sum ^{b}_{i=1} T_{\theta}(f_i, a_i) $$
$$\mathbb{E}_{\mathbb{P}^{(b)}_{F}\otimes \mathbb{P}^{(b)}_A}[e^{T_\theta}] =\frac{1}{b}\sum ^{b}_{i=1} e^{T_{\theta}(f_i, \Tilde{a}_i)} $$

where $b$ is the number of features in a batch $B$ whose mutual information to be computed with their corresponding attributes. We use a network with parameter set $\theta$ (see Fig. \ref{fig:template}) to compute the aforementioned arbitrary function $T_{\theta}(f_i, a_i)$ and $T_{\theta}(f_i, \Tilde{a}_i)$. By substituting $\mathbb{E}_{\mathbb{P}_{FA}}[T_\theta]$ and $\mathbb{E}_{\mathbb{P}_{F}\otimes \mathbb{P}_A}[e^{T_\theta}]$ in Eq. \ref{eq:mi} , the function $\mathcal{V}(\theta)$ is computed as:
\begin{equation}
     \mathcal{V}(\theta) = \frac{1}{b}\sum ^{b}_{i=1} T_{\theta}(f_i, a_i) - \text{log}(\frac{1}{b}\sum ^{b}_{i=1} e^{T_{\theta}(f_i, \Tilde{a}_i)})
\label{eq:nu}
\end{equation}
 As mentioned in \cite{belghazi2018mine}, the supremum of $\mathcal{V}(\theta)$ with respect to parameter set $\theta$ is a lower bound approximation of the mutual information between features $F$ and attributes $A$. Hence we use the following function $L$  as our objective function to train the network $N$.
\begin{equation}
    L(\theta)=-\mathcal{V}(\theta)
\label{eq:loss}
\end{equation}
\begin{equation}
    \nabla_\theta L(\theta) = - \Bigg( \mathbb{E}_{\mathbb{P}^{(b)}_{FA}} [\nabla T_\theta] - \frac{\mathbb{E}_{\mathbb{P}^{(b)}_{F} \otimes \mathbb{P}^{(b)}_{A}}[\nabla T_\theta e^{T_{\theta}}]}{\mathbb{E}_{\mathbb{P}^{(b)}_{F} \otimes \mathbb{P}^{(b)}_{A}}[e^{T_\theta}]}\Bigg)
\label{eq:grad}
\end{equation}
 At every training iteration, we use a different parameter set $\theta$ to compute the function $T_{\theta}$. As the training proceeds, the network minimizes (\ref{eq:loss}), thus maximizing $\mathcal{V}(\theta)$ with respect to $\theta$. This is equivalent to computing the supremum of $\mathcal{V}(\theta)$. Hence, the final value of $\mathcal{V}(\theta)$ at convergence is an approximation of Expressivity. (Eq. \ref{eq:mi})

 In equations (\ref{eq:nu}) and (\ref{eq:loss}), the objective function (and the lower bound of mutual information) is computed in a batch and not on the given set of features and their attribute values, thus making the gradients biased towards the minibatch, rather than the full batch. This issue has been identified in \cite{belghazi2018mine}, and can be mitigated by replacing the expectation term in the denominator of gradient update (\ref{eq:grad}) by an exponential moving average. More theoretical details related to this approximation of mutual information is provided in \cite{belghazi2018mine}.

\section{Proposed approach}
\label{sec:approach}
\newcommand{\tabrglayer}{
\begin{tabular}{cc}
\toprule
Layer & Description\\
\midrule
\texttt{Res4a2b} & 2\textsuperscript{nd} convolutional layer of 7\textsuperscript{th} Res-block \\
\texttt{Res5a2c} & 3\textsuperscript{rd} convolutional layer of 9\textsuperscript{th} Res-block\\
\texttt{Pool5}  & \thead{Final pooling layer which takes in the output of the\\ 11\textsuperscript{th} Res-block \texttt{Res\_5c}}\\
\texttt{FC-L2S} 	& \thead{Final fully connected which takes in the output of \texttt{Pool5} \\and computes $L_2$ softmax activation} \\

\bottomrule
\end{tabular}}
\newcommand{\tabaglayer}{
 \begin{tabular}{cc}
 \toprule
 Layer & Description\\
\midrule
\texttt{a9\_concat} & Concatenation layer of 10\textsuperscript{th} inception block \\
\texttt{b5\_concat} & Concatenation layer of 17\textsuperscript{th} inception block \\
\texttt{c3\_concat} & Concatenation layer of 36\textsuperscript{th} inception block \\
\texttt{Pool8x8}  & \thead{Final pooling layer which takes in the output of the\\ convolutional layer after 42\textsuperscript{nd} inception block \texttt{c10}}\\
\texttt{FC-L2S} 	& \thead{Final fully connected which takes in the output of\\ \texttt{Pool8x8} and computes $L_2$ softmax activation}\\
 \bottomrule
 \end{tabular}}
\begin{table}%
  \centering
  \subfloat[][Network A]{\tabrglayer}%
  \qquad
  \subfloat[][Network B]{\tabaglayer}
  \caption{Brief descriptions of the layers used to compute expressivity in both the networks. More architectural details are provided in \cite{ranjan2019fast}.}
  \label{tab:des_layers}
\end{table}
\begin{figure}
{\centering
\subfloat[]{\includegraphics[width = 0.5\linewidth, height = 40mm]{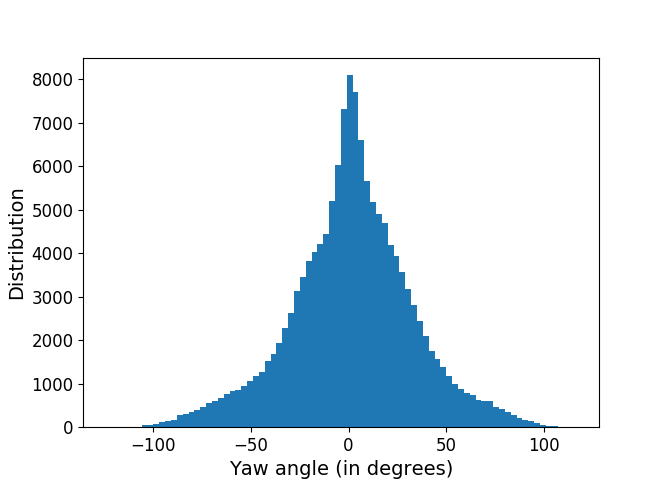}}
\subfloat[]{\includegraphics[width = 0.5\linewidth, height = 40mm]{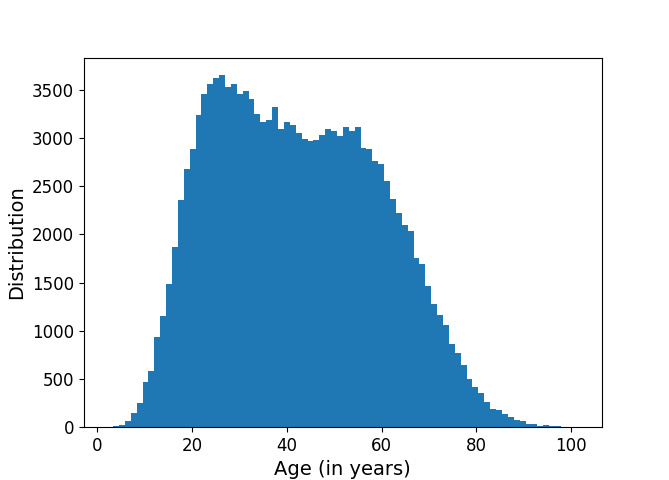}}
\caption{\small IJB-C dataset \cite{maze2018iarpa} shows enough variation with respect to (a.) Yaw, (b.) Age, which is required to compute expressivity of age and yaw, in a given set of IJB-C features.}
\label{fig:dis}
}
\end{figure} 
\subsection{Networks and datasets used}
To compute the expressivity of identity or attribute $A$ in a given set of features $F$, we extract the features $F$ using the following networks :\\
(1) \textbf{Network A} (Resnet-101 architecture) : The architecture is described in \cite{ranjan2019fast}. For investigating hierarchical course of the feedforward pass, we use a version of this network trained on a combined dataset of all the MS-Celeb-1M and UMD Faces images. For this trained network, we compute expressivity of attributes using features from these layers : \verb|Res4a2b, Res5a2c, Pool5, FC-L2S|.\\
(2) \textbf{Network B} (Inception Resnet v2 architecture).  The architecture is described in \cite{ranjan2019fast}. The training dataset of this network is same as that of Network A. We explore the following layers in the network: \verb|a9_concat, b5_concat, c3_concat, Pool8x8|, \verb|FC-L2S|.\\
Layer-wise details for both the networks are provided in Table \ref{tab:des_layers}.
For computing expressivity, we use IJB-C \cite{maze2018iarpa} as our test dataset. IJB-C dataset consists of 3531 identities with a total of 31,334 still images and 117,542 video frames collected in unconstrained settings. We extract the IJB-C features $F$ from different layers of aforementioned trained networks.  For computing the corresponding attributes $A$, we use the All-in-one network \cite{ranjan2017all}. 
\subsection{Attributes used}
 We compute the expressivity of identity and three attributes : yaw, sex, age in the extracted IJB-C features. To compute the yaw, sex and age of the corresponding IJB-C images, we use the All-in-one deep CNN proposed in \cite{ranjan2017all}, which simultaneously performs face detection, landmarks localization, pose estimation, sex recognition, smile detection, age estimation and face identification and verification. In Figure \ref{fig:dis}, we verify that the IJB-C datasets show enough variation with respect to yaw and age, so that we can insure that expressivity (which is a lower bound estimate of mutual information) is an accurate model for the corresponding attributes. The IJB-C dataset consists of 2203 male and 1328 female identities, which ensures that sufficient sex variation exists. 
 
 We now briefly explain the attribute vector $A$ introduced in Section \ref{sec:expressivity}. When considering sex, the vector $A$ consists of probability values of the sex being male (\verb|PR_MALE|), which are the outputs of All-in-One network in \cite{ranjan2017all}. When considering identity, $A$ is a discrete vector, where each element is a numerical identity label. We generate the attribute vector $A$ for yaw (in degrees) and age (in years) in a similar manner. The exact methodology to compute the expressivity using the feature descriptors and their respective attribute vector is provided in the next subsection.
\begin{algorithm}[H]
\floatname{algorithm}{Protocol}
\caption{Computing expressivity using flattened features}
\label{protocol1}
\begin{algorithmic}[1]
\STATE \textbf{Input}: layer $L$,  
\STATE \textbf{Input}: Set of $n$ images $I$
\STATE \textbf{Input}: attribute vector $A\in \mathbb{R}^{n \times 1}$
\STATE \textbf{Initialize} $E = []$
\STATE For a given image $i \in I$, extract feature $f_i$
\STATE Augmentation step $X = [F | A]$, where $F=[f_1,f_2 \ldots f_n]^T$
\FOR {$iter:$ 1 to M}
\STATE Initialize MINE network according to dimensions of $X$
\STATE $E\leftarrow$MINE($X$)
\ENDFOR
\STATE \textbf{return} Expressivity = Average($E$)
\end{algorithmic}
\end{algorithm} 
\begin{algorithm}[H]
\floatname{algorithm}{Protocol}
\caption{Computing expressivity using unflattened feature maps}
\label{protocol2}
\begin{algorithmic}[1]
\STATE \textbf{Input}: layer $L$, with $k$ channels, each of dimension $d\times d$
\STATE \textbf{Input}: Set of $n$ images $I$
\STATE \textbf{Input}: attribute vector $A\in \mathbb{R}^{n \times 1}$
\STATE \textbf{Initialize} $E = []$
\STATE $S=$ Subset of randomly selected $z$ (out of $k$) channels 
\STATE For a given image $i$, vectorize and concatenate all $z$ maps in $S$, to generate vector $f_i$ of dimension $m\times1$, where $m=d*d*z$
\STATE Augmentation step $X = [F | A]$ where $F=[f_1,f_2 \ldots f_n]^T$
\FOR {$iter:$ 1 to M}
\STATE Initialize MINE with input dimensions of $X \in \mathbb{R}^{n\times m+1}$
\STATE $E\leftarrow$MINE($X$)
\ENDFOR
\STATE \textbf{return} Expressivity = Average($E$)
\end{algorithmic}
\end{algorithm}
\subsection{Protocols to compute expressivity}
\label{subsec:proto}
\vspace{-0.25cm}
In this work, we define expressivity as a lower bound approximation of MINE, as explained in section \ref{sec:expressivity}. We consider two protocols (described above): Protocol-1, to compute the expressivity of attributes in flattened features (i.e. features from fully connected or pooling layers which are extracted as vectors). Protocol-2, to compute the expressivity of attributes in unflattened feature maps (i.e. feature maps from a convolutional layer which are extracted as 2D channels). For both of these protocols, we need a set of $n$ images, and their corresponding attributes $A\in \mathbb{R}^{n\times 1}$. In step 8 of Protocol-1 and step 9 of Protocol-2, we initialize a MINE approximation network according to the input dimension of the augmented matrix. As explained in Section \ref{sec:expressivity}, we train the network to compute the lower bound approximation of mutual information between features $F$ and attribute $A$. We use a simple multi layer perceptron (MLP) network, described in Figure \ref{fig:template}, for computing $T_\theta$ in (\ref{eq:loss}). The network consists of two hidden layers with 256 and 128 units. These layers are followed by ELU activations. We use this architecture consistent throughout our experiments. When using different sets of features for $F$, the only architectural changes in the network are made in the input layer dimension, according to the feature dimension. The network is trained until the loss in (\ref{eq:loss}) converges and expressivity is computed using the protocols. In step 8 in Protocol-1 and step 9 in Protocol-2 we initialize and train the MINE network multiple times ($M$) to increase number parameter sets $\theta$ in (\ref{eq:mi}), among which the supremum is to be found. In all our experiments, we use $M=16$. Also, in step 5 of Protocol-2 we select only a subset of features maps (channels), as mentioned in \cite{alain2016understanding}. Note that, for any attribute vector $A$ (yaw/identity/age/sex), we use the same feature subset $S$.

Apart from analyzing the network layer features, we also investigate the presence of various attributes in the raw RGB face image in Section \ref{sec:experiments}. For this, we average the R,G and B channels to generate a grayscale image. Following this, we vectorize the image and use it as feature $f_i$ in Protocol-1.
\subsection{Training linear classifiers}
To verify that expressivity correctly models the information of attributes in features, we show its correlation with error-rates of linear classifiers trained on the corresponding features. We randomly select a subset of 5000 IJB-C images and extract their features. To train the linear classifier we use 3000 features and test it on 2000 features. This is a trivial task for flattened features. However, to compute the error rate in feature maps from higher layers of the network, we use the same subset $S$ of feature maps as selected in step 5 of Protocol-2. Following this, we vectorize and concatenate them as in Step 7. We provide more specific details in the next section.

\section{Experiments}
\label{sec:experiments}
\newcommand{\tabrg}{
\begin{tabular}{ccccc}
\toprule
Layer & Dim. ($c\times d\times d$)& Protocol & $z$ & Feat. dim\\
\midrule
\texttt{Res4a2b} &$1024\times14\times14$&2&11&2156\\
\texttt{Res5a2c}  &$2048\times7\times7$&2&42&2057\\
\texttt{Pool5}  &2048	&1&-&2048\\
\texttt{FC-L2S} 	&512	&1&-&512\\

\bottomrule
\end{tabular}}
\newcommand{\tabag}{
 \begin{tabular}{ccccc}
 \toprule
 Layer & Dim. ($c\times d\times d$)& Protocol & $z$ & Feat. dim\\
 \midrule
 \texttt{a9\_concat}&$128\times35\times35$&2&3&3675\\
 \texttt{b5\_concat}  &$384\times17\times17$&2&13&3757\\
 \texttt{c3\_concat}  &$128\times8\times8$&2&59&3776\\
 \texttt{Pool8x8}  &1536	&1&-&1536\\
 \texttt{FC-L2S}	&512	&1&-&512\\
 \bottomrule
 \end{tabular}}
 
\begin{table}%
  \centering
  \subfloat[][Network A]{\tabrg}%
  \qquad
  \subfloat[][Network B]{\tabag}
  \caption{\small Network A and B layers used to compute expressivity. $c$ : number of channels in a given layer, $d$: channel dimension, $z$ : Number of channels selected out of $c$ channels, to generate a subset of features}%
  \label{tab:layers}%
\end{table}
Using the protocols described in Section \ref{sec:approach} and the expressivity measure defined in \ref{sec:expressivity}, we extract features from different layers of Networks A and B and use them to train a network (Figure \ref{fig:template}) to compute expressivity of various attributes.
\subsection{Hierarchical course of feedforward pass}
\label{subsec:hierar}
\begin{figure*}
   \centering
   \subfloat[][Network 
   A]{\includegraphics[width=3.25in]{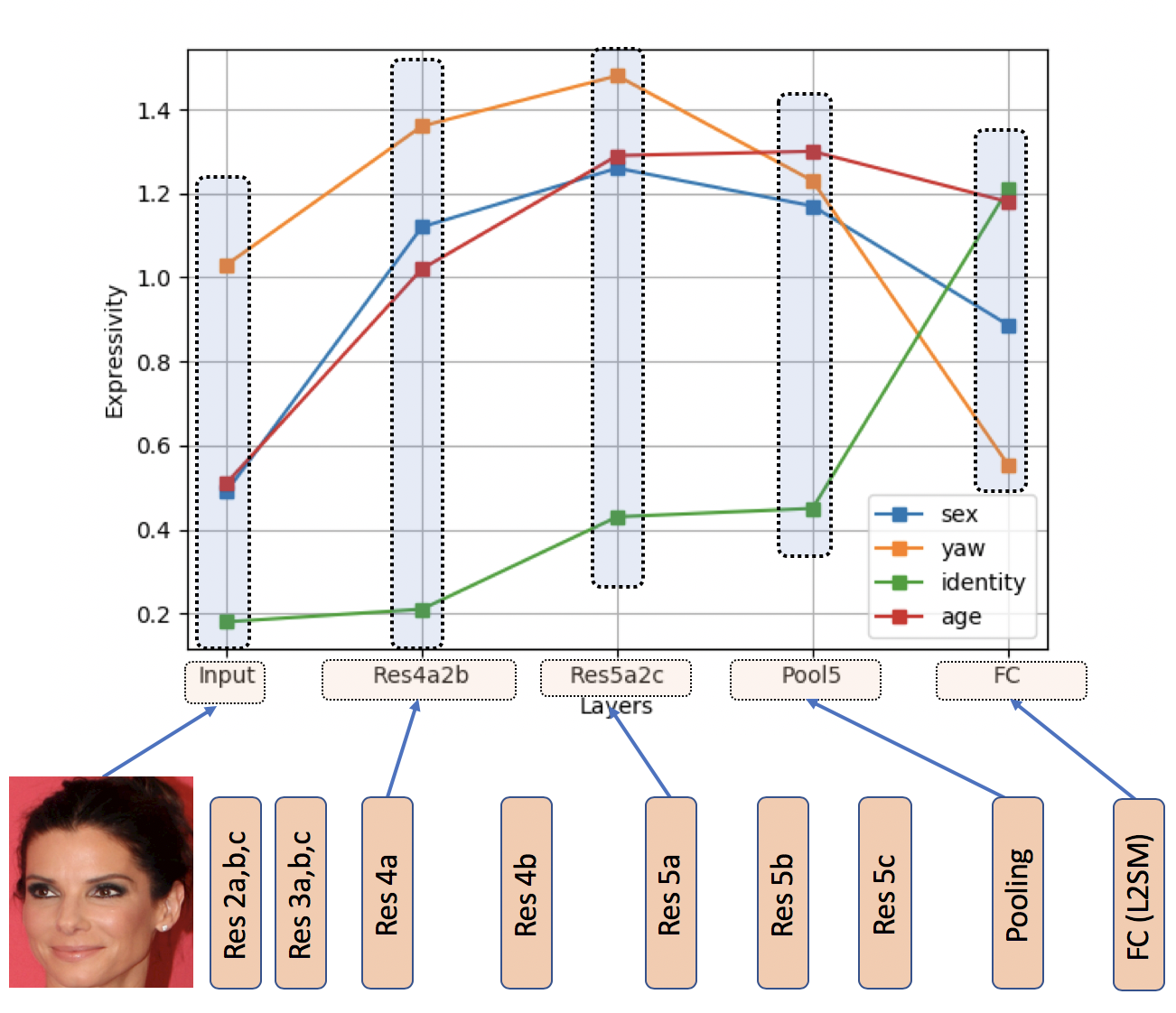}}
   \rulesep
    \subfloat[][Network 
    B]{\includegraphics[width=3.10in]{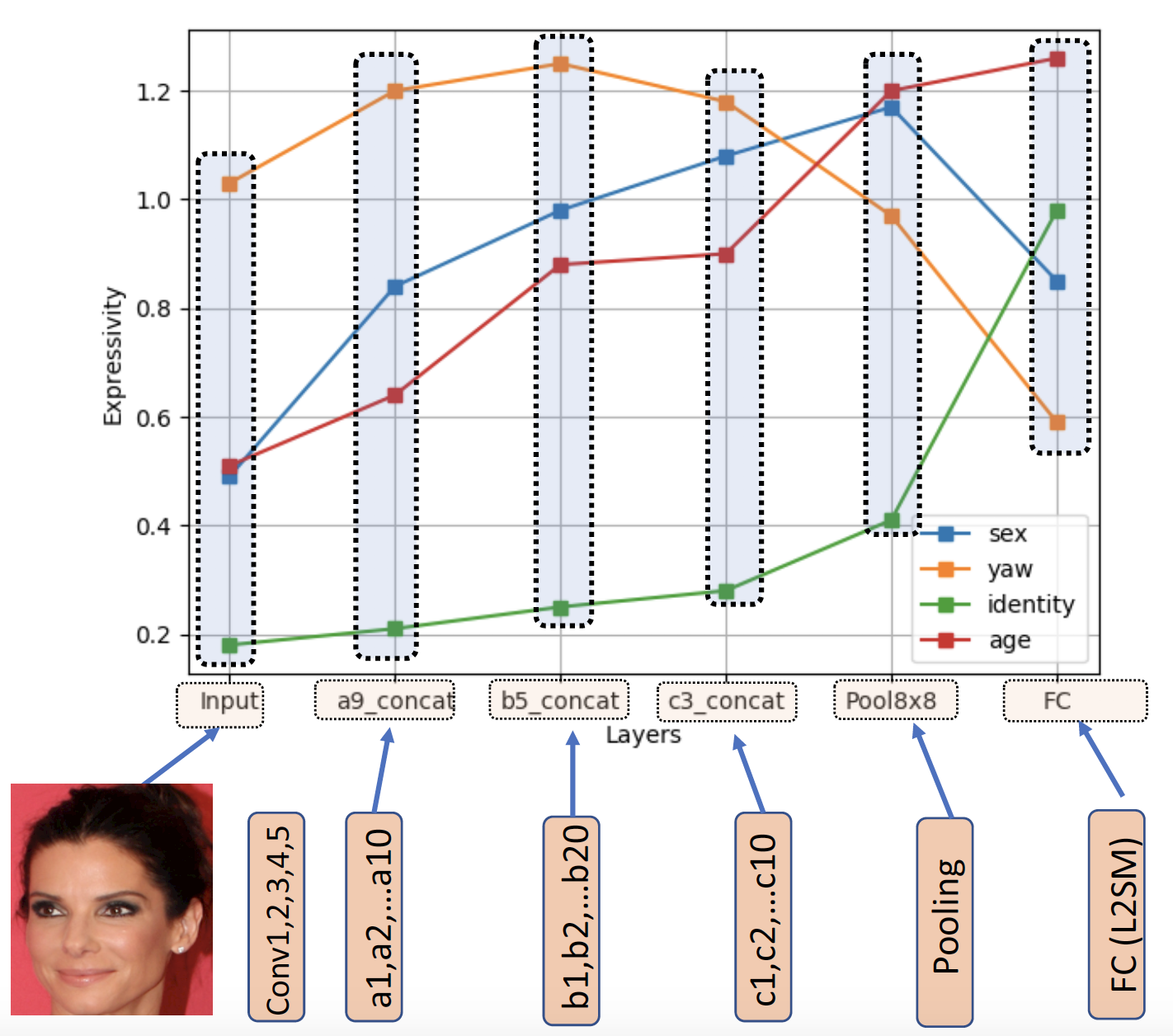}}
    \caption{\small Expressivity of identity, age, sex and yaw in input image and layerwise features from both the networks. The source image for face in this figure is attributed to Eva Rinaldi under the
\href{https://creativecommons.org/licenses/by-sa/2.0/deed.en}{[cc-by-sa-2.0]} creative commons licenses respectively. The face was cropped from the source image.}
    \label{fig:hierar}
\end{figure*}

 Table \ref{tab:layers} shows the network layers explored, along with the final dimension of the features used for computing expressivity for both the networks. The layerwise expressivity values for Networks A and B are shown in Figure \ref{fig:hierar}. It should be noted that both the networks were trained using identity-supervision and no supervision based on pose, sex and age. Our inference is listed as follows:
\begin{itemize}[leftmargin=*]
    \item In both networks A and B, we find that the expressivity of yaw, sex and age is high and that of identity is the lowest in the shallower layers (\verb|Res4a2b, Res5a2c| in Network A; \verb|a9_concat, b5_concat, c3_concat| in Network B) and input image. This shows that yaw and sex are high level face features as compared to identity, which cannot be extracted using shallow layers.
    \item As we examine the deepest layer (FC-L2S), the expressivity of yaw and sex attain their lowest values, whereas identity and age have very high expressivity. This shows that identity and age are more fine grained features compared to other attributes.
    \item There is a rapid increase in the identity expressivity from the pooling to fully connected layers, in both the networks.
    \item Comparing the expressivity values of all attributes except identity in the final layer, we can infer that for identity recognition, yaw is the least important and age is the most important attribute.
\end{itemize}

There are three reasons for the relatively high expressivity of the
age when compared to the other attributes. 

First, in the IJB-C dataset most of the images were acquired over a short
period of time, therefore, it can happen that a given age correlates with identity. Second, we used an automated algorithm \cite{ranjan2017all} to estimate the age, and therefore it is computed from the appearance of the face. All attributes were computed automatically, but we can expect that in relative terms the age is the least accurate of all attributes automatically computed. Third, the entropy of the age (Fig. \ref{fig:dis}(b)) is higher than the entropy of the other attributes and this could increase the mutual information component
of the expressivity.

\textbf{Discussion of the data processing inequality}: The data processing inequality (DPI) \cite{cover2012elements} states that for three random variables $P, Q, R$ forming a Markov chain $P\longrightarrow Q\longrightarrow R$, $$\text{MI}(P,Q)\geq \text{MI}(P,R). $$

The data processing inequality formalizes the concept that no processing of data can increase mutual information. To make this more concrete, let $P$ be any random variable (e.g. sex, yaw, identity), and let $Q,R$ be features for different layers in a network where $R$ is a deterministic function of $Q$, i.e. $R$ is deeper than $Q$. Since $R$ is a function of $Q$ then $P,Q,R$ forms a Markov chain \cite{cover2012elements}. It follows from the DPI that the information about $P$ contained in the features cannot increase as we go deeper. 


The expressivity results in Figure \ref{fig:hierar} are not monotonically decreasing, which might seem like a contradiction to DPI. However, as pointed out in \cite{alain2016understanding}, the features in our context denote \textit{representation}, rather than \textit{information content} described in Information Theory. \textit{Representations} are more closely related to predictability of a specific attribute, as compared to information-theoretic content. Hence, in this work, expressivity refers to the accord between and attribute and attribute, rather than its theoretical information content.


\begin{table}
\centering
\begin{tabular}{ccc}
\toprule
Layer  & \thead{Yaw \\regress\textsuperscript{n} error} & \thead{Yaw \\expressivity}\\
\midrule
\texttt{Res4a2b} &11.42&1.36\\
\texttt{Res5a2c}  &\textcolor{green}{8.64}&\textcolor{green}{1.48}\\
\texttt{Pool5}  &11.57&1.23 \\
\texttt{FC-L2S} 	&\textcolor{red}{11.65}&\textcolor{red}{0.59}\\
\bottomrule
\end{tabular}
\vspace{.5em}
\caption{\small Comparison  of yaw regression errors with their corresponding expressivity values, in different layers of Network A. The highest accuracy (or lowest error) corresponds to highest expressivity and lowest accuracy (or highest error) corresponds to lowest expressivity.}
\label{table:rg1_compare}
\end{table}
\textbf{Relation with linear separability}: Alain and Bengio \cite{alain2016understanding} show that the linear separability of features with respect to output classes, which provided supervision, monotonically increases as we go deeper into the network. From Figure \ref{fig:hierar}, we find this to be true for expressivity identity as well, which provided supervision during training . \\

In order to ensure that the expressivity values correlate with feature vectors, we compute the accuracy/error rate obtained by training a linear classifier and testing it directly using features from the aforementioned layers in Network A, as explained Section \ref{sec:approach}. To analyze the yaw expressivity values, we first train a simple linear regression model on 3000 randomly selected features (extracted from IJB-C images) and evaluate its regression error on 2000 IJB-C features.The corresponding results are presented in Table \ref{table:rg1_compare}, from which we can infer that expressivity values do correlate with regression errors for yaw.
\vspace{-0.2cm}
\subsection{Temporal course of training}
\begin{figure*}
    \centering
    \subfloat[]{\includegraphics[width=0.515\linewidth]{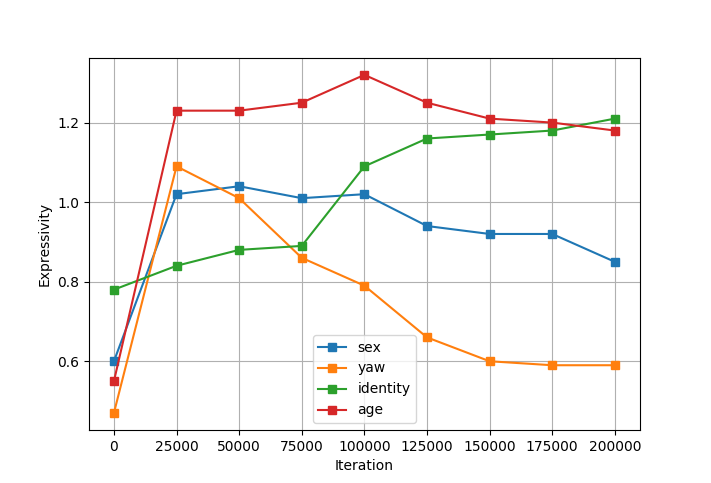}}
    \subfloat[]{\includegraphics[width=0.495\linewidth]{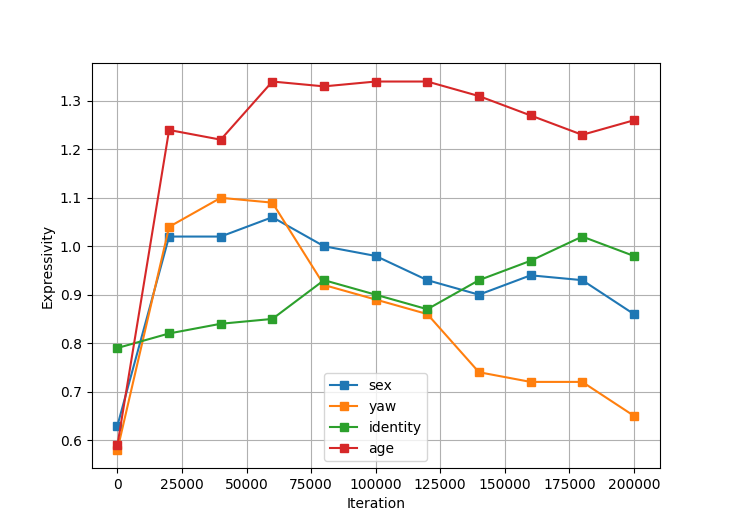}}
    \caption{\small Expressivity of identity, age, sex and yaw in final layer (FC-L2S) features of a.) Network A and b.) Network B.  Decreasing expressivity of task irrelevant attributes (yaw, age, sex) is a part of training. Observed rate of decrease : Age $<$ Sex $<$ Yaw }
    \label{fig:temporal}
\end{figure*}
We also analyze the training process of Network A and B and investigate the changes in the expressivity of yaw, sex, identity and age in the final layer (FC-L2S) of these networks with respect to its training iterations. The features at all iterations ($>0$) are 512 dimensional and are flattened and Protocol-1 is used for computing attribute-wise expressivity, along with specifications for FC-L2S mentioned in Table \ref{tab:layers}. The features at iteration 0, represent the final layer features of the networks trained on ImageNet \cite{Imagenet}, without the final fully connected layer for identity recognition. These features are therefore 2048 and 1536 dimensional for networks A and B respectively. The results are presented in Figure \ref{fig:temporal}.  Our observations are listed below:
\begin{itemize}[leftmargin=*]
    \item We find that the expressivity of yaw, age and sex reach their peak values in the first $~$25000 iterations for Network A and $~$40000 iterations for Network B, to learn the general concept of facial pose, age and sex.
    \item Following that, we find that the yaw expressivity decreases rapidly as the training proceeds, showing that making features almost agnostic to pose variance is an essential part of the training process. 
    \item For both networks A and B, the expressivity of age and sex decreases slightly after their corresponding expressivity peaks are attained, during the course of training. However, compared to yaw, the rate of decay in the expressivity of age and sex is low. This shows that age and sex are more important for identity recognition than face yaw.
    \item Observing the expressivity values in the final iteration, we can infer that for identity recognition, the following is the order of relevance of attributes for which the network does not receive any supervision : Age $>$ Sex $>$ Yaw. The opposite of this order is observed in the rate by which the expressivity values of yaw, age and sex decreases, i.e. the rate of decrease is : Age $<$ Sex $<$ Yaw. This is true for both networks A and B.
    \item The expressivity of identity generally increases during training for both networks A and B.
    \item Features extracted from the final layer of Iteration 0 model (Imagenet features), express identity better than other attributes. This is because the Imagenet features express `objectness', which is closely related to identity as compared to other attributes (yaw, age, sex).  
\end{itemize}
Similar to what we did in Section \ref{subsec:hierar}, we compare the expressivity values to the corresponding error rates by training and testing linear classifiers directly on the final fully connected layer features. The results are presented in Table \ref{table:rg1_temporal_compare}, where we again find that there exists correlation between expressivity values  and age/yaw regression errors. 
\begin{table}
\centering
\begin{tabular}{ccc|cc}
\toprule
Iteration & \thead{Yaw \\ error} & \thead{Yaw \\expr.}&\thead{Age \\ error} & \thead{Age \\expr.}\\
\midrule
$T_1$ &\textcolor{green}{9.40}&\textcolor{green}{1.13}&8.06&1.22\\
$T_2$  &11.27&0.68&\textcolor{green}{7.70}&\textcolor{green}{1.39}\\
$T_3$  &\textcolor{red}{11.65}&\textcolor{red}{0.59}&\textcolor{red}{8.11}&\textcolor{red}{1.19}\\
\bottomrule
\end{tabular}
\caption{\small Comparison of  age and yaw regression errors with their corresponding expressivity values in 3 iterations $T_1,T_2,T_3$. For yaw, $T_1,T_2,T_3=25k,100k,200k$ iterations.  For age $50k,100k,200k$}
\label{table:rg1_temporal_compare}
\end{table}
\subsection{Advantage of expressivity over other techniques}
%
The following are the advantages of Expressivity over existing interpretability techniques:\\
\textbf{Comparison of several attributes}: Comparing expressivity of different attributes in a given set of features is important for understanding the information organization in a trained network. As shown in Figures \ref{fig:temporal} and \ref{fig:hierar}, expressivity helps put the error rates of all attributes on the sample scale, thus enabling their comparisons. This cannot be achieved by directly using the accuracy/error rates of linear classifiers, as different attributes have different scales and evaluation metrics. \\
\textbf{Useful for any physical concept (discrete/continuous)}: Although we used expressivity to analyze face recognition networks in terms of facial attributes, it can be used for a network with respect to any physical attribute. For instance, we can compute the expressivity of `stripes' concept in a set of features $F$ if we have a binary attribute vector $A$, denoting the presence or absence of `stripes'. This is similar to TCAVs \cite{kim2017interpretability}. However, TCAVs cannot be directly used to quantify the content of continuous concept (like pose angle), since we need images with which demonstrate absence of that concept to train CAVs, and this is not trivial for concepts like pose angle, age etc.  \\
\textbf{No dependence on training classes}:  Methods like \cite{kim2017interpretability} and \cite{selvaraju2017grad} require computing change of logit values. However, for images not in training classes, this value is not meaningful. Using expressivity allows to measure information about attributes that were not explicitly included in  training.  \\
\vspace{-0.5cm}
\subsection{Relation with information bottleneck theory}
Saxe et al \cite{saxe2018information} provide a response to the Information Bottleneck theory \cite{shwartz2017opening}, and claim that when an input domain consists of a subset of task-relevant and task-irrelevant information, hidden representations do compress the task-irrelevant information. In our context, identity is the task-relevant information, since the networks only receive identity supervision. The task-irrelevant information includes yaw, age and sex information. In our work, we concur with the claims of \cite{saxe2018information}. It can be seen in Figures \ref{fig:temporal} the expressivity of task-irrelevant attributes (yaw, age, sex) in the final layer decreases as we train the network, depicting the `compression' phase. This occurs while the identity expressivity increases as the training progresses, which corresponds to the `fitting' phase of the network. Hence, we also verify another result in \cite{saxe2018information}, that this compression happens concurrently with the fitting process rather than during a subsequent compression period.
\section{Discussion and future work}
We present an approach to quantify the information learned by a face recognition network about several attributes and identities, by computing their expressivity in a given set of features. The scale of this measure is agnostic to the attribute being examined.  We use this measure to analyse layer-wise features, and temporal snapshots of the final fully connected layer in two face recognition networks. We make some important observations in both the networks we investigated: (1) The mid-layer and shallow layer features effectively capture task-irrelevant information (about yaw, sex, age). The deeper layers encode task-relevant identity information. (2.) During the training process, the expressivity of identity increases while that of yaw, sex and age decreases, thus showing that \textit{decreasing expressivity of task-irrelevant attributes is a part of learning}. Expressivity of yaw, especially, decreases very rapidly. (3.) Using the expressivity values in the final layer of trained networks, we find the following order of attribute-wise relevance for identity recognition : Age $>$ Sex $>$ Yaw. This is opposite to the order of the rate by which expressivity of these three attributes decrease during training. We also relate our findings with existing works on interpretability and information bottleneck theory.

There are other face attributes, such as  facial expression, presence of eyeglasses, beard, hairstyle etc. which play a crucial role in identity recognition. One future avenue of research is to extend our work for these attributes  and obtain a more exhaustive ordering of face attributes in terms of their relevance to face recognition. Also, one could investigate the training processes of layers other than fully connected layers. Finally, one could estimate the expressivity of attributes in face descriptors extracted from other networks, like \cite{VGGFace} and \cite{Swami_2016_triplet}.

{\small
    \bibliographystyle{unsrt}
    \bibliography{biblio}

\begin{thebibliography}{10}

\bibitem{hill2019deepNatureMI}
M~Q Hill, C~J Parde, C~D Castillo, Y~I Colon, R~Ranjan, J-C Chen, V~Blanz, and
  A~J O'Toole.
\newblock Deep convolutional neural networks in the face of caricature:
  Identity and image revealed.
\newblock {\em Nature Mach Intel}, (in press).

\bibitem{DBLP:journals/corr/abs-1904-01219}
S~Nagpal, M~Singh, R~Singh, M~Vatsa, and N~K Ratha.
\newblock Deep learning for face recognition: Pride or prejudiced?
\newblock {\em CoRR}, abs/1904.01219, 2019.

\bibitem{parde2017face}
C~J Parde, C~D Castillo, M~Q Hill, Y~I Colon, S~Sankaranarayanan, Jun-Cheng
  Chen, and Alice~J O’Toole.
\newblock Face and image representation in deep cnn features.
\newblock In {\em 2017 12th IEEE International Conference on Automatic Face \&
  Gesture Recognition (FG 2017)}, pages 673--680. IEEE, 2017.

\bibitem{Givens:2013fk}
G.~H. Givens, J.~R. Beveridge, P.~J. Phillips, B.~A. Draper, Y.~M. Lui, and
  D.~S. Bolme.
\newblock Introduction to face recognition and evaluation of algorithm
  performance.
\newblock {\em Computational Statistics and Data Analysis}, 67:236--247, 2013.

\bibitem{Lee:2014dn}
Y.~Lee, P.~J. Phillips, J.~J. Filliben, J.~R. Beveridge, and H.~Zhang.
\newblock Generalizing face quality and factor measures to video.
\newblock In {\em International Joint Conference on Biometrics (IJCB)}, 2014.

\bibitem{ranjan2019fast}
R~Ranjan, A~Bansal, J~Zheng, H~Xu, J~Gleason, B~Lu, A~Nanduri, J-C Chen, C~D
  Castillo, and R~Chellappa.
\newblock A fast and accurate system for face detection, identification, and
  verification.
\newblock {\em IEEE Transactions on Biometrics, Behavior, and Identity
  Science}, 1(2):82--96, 2019.

\bibitem{bansal2018deep}
A~Bansal, R~Ranjan, C~D Castillo, and R~Chellappa.
\newblock Deep features for recognizing disguised faces in the wild.
\newblock In {\em 2018 IEEE/CVF Conference on Computer Vision and Pattern
  Recognition Workshops (CVPRW)}, pages 10--106. IEEE, 2018.

\bibitem{schroff2015facenet}
F~Schroff, D~Kalenichenko, and J~Philbin.
\newblock Facenet: A unified embedding for face recognition and clustering.
\newblock In {\em Proceedings of the IEEE Conference on Computer Vision and
  Pattern Recognition}, pages 815--823, 2015.

\bibitem{taigman2014deepface}
Y~Taigman, M~Yang, M~Ranzato, and L~Wolf.
\newblock Deepface: Closing the gap to human-level performance in face
  verification.
\newblock In {\em Proceedings of the IEEE Conference on Computer Vision and
  Pattern Recognition}, pages 1701--1708, 2014.

\bibitem{deng2018arcface}
J~Deng, J~Guo, and S~Zafeiriou.
\newblock Arcface: Additive angular margin loss for deep face recognition.
\newblock {\em preprint arXiv:1801.07698}, 2018.

\bibitem{yin2018towards}
B~Yin, L~Tran, H~Li, X~Shen, and X~Liu.
\newblock Towards interpretable face recognition.
\newblock {\em arXiv preprint arXiv:1805.00611}, 2018.

\bibitem{kim2014bayesian}
B~Kim, C~Rudin, and J~A Shah.
\newblock The bayesian case model: A generative approach for case-based
  reasoning and prototype classification.
\newblock In {\em Advances in Neural Information Processing Systems}, pages
  1952--1960, 2014.

\bibitem{kim2017interpretability}
B~Kim, M~Wattenberg, J~Gilmer, C~Cai, J~Wexler, F~Viegas, and R~Sayres.
\newblock Interpretability beyond feature attribution: Quantitative testing
  with concept activation vectors (tcav).
\newblock {\em preprint arXiv:1711.11279}, 2017.

\bibitem{alain2016understanding}
G~Alain and Y~Bengio.
\newblock Understanding intermediate layers using linear classifier probes.
\newblock {\em preprint arXiv:1610.01644}, 2016.

\bibitem{koh2017understanding}
P~W Koh and P~Liang.
\newblock Understanding black-box predictions via influence functions.
\newblock In {\em Proceedings of the 34th International Conference on Machine
  Learning-Volume 70}, pages 1885--1894. JMLR. org, 2017.

\bibitem{selvaraju2017grad}
R~R Selvaraju, M~Cogswell, A~Das, R~Vedantam, D~Parikh, and D~Batra.
\newblock Grad-cam: Visual explanations from deep networks via gradient-based
  localization.
\newblock In {\em Proceedings of the IEEE International Conference on Computer
  Vision}, pages 618--626, 2017.

\bibitem{chattopadhay2018grad}
A~Chattopadhay, A~Sarkar, P~Howlader, and V~N Balasubramanian.
\newblock Grad-cam++: Generalized gradient-based visual explanations for deep
  convolutional networks.
\newblock In {\em 2018 IEEE Winter Conference on Applications of Computer
  Vision (WACV)}, pages 839--847. IEEE, 2018.

\bibitem{hill2018deep}
M~Q Hill, C~J Parde, C~D Castillo, Y~I Colon, R~Ranjan, JC~Chen, V~Blanz, and
  A~J O'Toole.
\newblock Deep convolutional neural networks in the face of caricature:
  Identity and image revealed.
\newblock {\em arXiv preprint arXiv:1812.10902}, 2018.

\bibitem{tishby2015deep}
N~Tishby and N~Zaslavsky.
\newblock Deep learning and the information bottleneck principle.
\newblock In {\em 2015 IEEE Information Theory Workshop (ITW)}, pages 1--5.
  IEEE, 2015.

\bibitem{belghazi2018mine}
M~I Belghazi, A~Baratin, S~Rajeswar, S~Ozair, Y~Bengio, A~Courville, and R~D
  Hjelm.
\newblock Mine: mutual information neural estimation.
\newblock {\em arXiv preprint arXiv:1801.04062}, 2018.

\bibitem{maze2018iarpa}
Brianna Maze, Jocelyn Adams, James~A Duncan, Nathan Kalka, Tim Miller, Charles
  Otto, Anil~K Jain, W~Tyler Niggel, Janet Anderson, Jordan Cheney, et~al.
\newblock Iarpa janus benchmark-c: Face dataset and protocol.
\newblock In {\em 2018 International Conference on Biometrics (ICB)}, pages
  158--165. IEEE, 2018.

\bibitem{ranjan2017all}
R~Ranjan, S~Sankaranarayanan, C~D Castillo, and R~Chellappa.
\newblock An all-in-one convolutional neural network for face analysis.
\newblock In {\em 2017 12th IEEE International Conference on Automatic Face \&
  Gesture Recognition (FG 2017)}, pages 17--24. IEEE, 2017.

\bibitem{cover2012elements}
T~M Cover and J~A Thomas.
\newblock {\em Elements of information theory}.
\newblock John Wiley \& Sons, 2012.

\bibitem{Imagenet}
J.~Deng, W.~Dong, R.~Socher, L.-J. Li, K.~Li, and F.-F. Li.
\newblock Imagenet: A large-scale hierarchical image database.
\newblock In {\em IEEE Conference on Computer Vision and Pattern Recognition},
  pages 248--255, June 2009.

\bibitem{saxe2018information}
A~M Saxe, Y~Bansal, J~Dapello, M~Advani, A~Kolchinsky, B~D Tracey, and D~D Cox.
\newblock On the information bottleneck theory of deep learning.
\newblock 2018.

\bibitem{shwartz2017opening}
R~Shwartz-Ziv and N~Tishby.
\newblock Opening the black box of deep neural networks via information.
\newblock {\em arXiv preprint arXiv:1703.00810}, 2017.

\bibitem{VGGFace}
O~M Parkhi, A~Vedaldi, and A~Zisserman.
\newblock Deep face recognition.
\newblock In {\em BMVC}, volume~1, page~6, 2015.

\bibitem{Swami_2016_triplet}
S~Sankaranarayanan, A~Alavi, C~D Castillo, and R~Chellappa.
\newblock Triplet probabilistic embedding for face verification and clustering.
\newblock In {\em 2016 IEEE 8th International Conference on Biometrics Theory,
  Applications and Systems (BTAS)}, 2016.

\end{thebibliography}
}

\end{document}